# Classification of Psychiatry Clinical Notes by Diagnosis: A Deep Learning and Machine Learning Approach

**Sergio Rubio-Martín**[1], **María Teresa García-Ordás**[1], **Antonio Serrano-García**[2], **Clara Margarita Franch-Pato**[2], **Arturo Crespo-Álvaro**[1], **and José Alberto Benítez-Andrades**[1]

[1]**ALBA Research Group, Department of Electric, Systems and Automatics Engineering, Universidad de León, Campus of Vegazana s/n, León, 24071, León, Spain**
[2]**Servicio de Psiquiatría, Complejo Asistencial Universitario de León (CAULE), 24008, León, Spain**

Corresponding author:
Sergio Rubio-Martín[1]

Email address: srubm@unileon.es

## ABSTRACT

The classification of clinical notes into specific diagnostic categories is critical in healthcare, especially for mental health conditions like Anxiety and Adjustment Disorder. In this study, we compare the performance of various Artificial Intelligence models, including both traditional Machine Learning approaches (Random Forest, Support Vector Machine, K-nearest neighbors, Decision Tree, and eXtreme Gradient Boost) and Deep Learning models (DistilBERT and SciBERT), to classify clinical notes into these two diagnoses. Additionally, we implemented three oversampling strategies: No Oversampling, Random Oversampling, and Synthetic Minority Over-sampling Technique (SMOTE), to assess their impact on model performance. Hyperparameter tuning was also applied to optimize model accuracy. Our results indicate that oversampling techniques had minimal impact on model performance overall. The only exception was SMOTE, which showed a positive effect specifically with BERT-based models. However, hyperparameter optimization significantly improved accuracy across the models, enhancing their ability to generalize and perform on the dataset. The Decision Tree and eXtreme Gradient Boost models achieved the highest accuracy among machine learning approaches, both reaching 96%, while the DistilBERT and SciBERT models also attained 96% accuracy in the deep learning category. These findings underscore the importance of hyperparameter tuning in maximizing model performance. This study contributes to the ongoing research on AI-assisted diagnostic tools in mental health by providing insights into the efficacy of different model architectures and data balancing methods.

## 1 INTRODUCTION

The field of medicine has advanced rapidly in recent decades due to technological innovations that have transformed both the diagnostic and treatment phases. In the mental health sector, particularly in psychiatry, there has been a paradigm shift, with a growing focus on understanding the brain and the underlying mechanisms that regulate behavior, emotions, and responses to external or internal changes. Despite these advances, psychiatric diagnosis still faces significant challenges due to the subjective and complex nature of the symptoms presented by patients, as well as the frequent overlap between different mental disorders. Moreover, despite the progress made in the field of medicine, the number of patients suffering from mental disorders has not decreased but has instead increased since 2019, when 970 million people were living with a mental disorder (World Health Organization, 2022a). The World Health Organization (WHO) is concerned not only about these numbers but also about the increase in mental disorders diagnosed during the COVID-19 pandemic, with cases of anxiety disorder rising by an estimated 26% (World Health Organization, 2022b).

In the field of mental healthcare, two major disciplines coexist: psychology and psychiatry. Although they share the common goal of improving mental well-being, they differ significantly in their training, methods, and approaches to treatment. Psychology is the scientific study of mental processes and behavior, including both internal mental activities, such as thoughts and emotions, and externally observable behaviors (Henriques and Michalski, 2020). Psychological practice primarily involves therapeutic methods based on dialogue and behavioral interventions, such as cognitive-behavioral therapy, humanistic therapy, or psychodynamic approaches. These treatments focus on modifying dysfunctional behaviors, emotions, and thoughts, typically following a non-medical model.

Psychiatry, by contrast, is a branch of medicine concerned with the diagnosis, treatment, and prevention of mental, emotional, and behavioral disorders. Psychiatrists, as medical doctors, are trained to assess both psychological symptoms and their biological underpinnings. They can prescribe pharmacological treatments and often manage complex cases involving severe mental illnesses, such as schizophrenia, bipolar disorder, major depression, or severe anxiety disorders (Kendler et al., 2011).

While psychology and psychiatry approach mental health from different perspectives—psychology focusing more on psychological and social aspects, psychiatry integrating biological, psychological, and pharmacological considerations—the two disciplines are complementary and increasingly collaborate in interdisciplinary mental health teams to provide holistic patient care.

While psychological conditions often involve significant distress, psychiatric disorders may pose more serious risks to patients' lives, including an increased risk of suicide. Moreover, it has been extensively documented that individuals suffering from severe mental disorders frequently experience reduced life expectancy. For example, people diagnosed with schizophrenia have an estimated life expectancy that is 10 to 20 years shorter than that of the general population (Nimavat et al., 2023). In addition, many individuals with mental illnesses face a substantial treatment gap, with only 29% of those with psychosis and 33% of those with depression receiving formal mental health care (World Health Organization, 2021; Moitra et al., 2022). These challenges highlight the pressing need for innovative approaches to support mental healthcare systems and improve access and quality of care.

One of the biggest problems for people suffering from a mental illness when seeking help from a public psychologist or psychiatrist is the long waiting time to get an appointment. Timely access to professionals would help patients receive a diagnosis and appropriate treatment for their condition. However, due to the lack of human and economic resources, as well as the time required to get an appointment, we propose a solution aimed at reducing the workload in the classification of clinical notes, also known as electronic health records (EHR).

It has been shown that artificial intelligence (AI) models have helped in various medical fields. For example, in oncology, AI has become a valuable tool for predicting cancer (Liu et al., 2020; Alanazi, 2023; Briganti and Le Moine, 2020). Additionally, AI continues to be useful in predicting cancer recurrence (Zhang et al., 2023). Some AI models based on images have been used to detect different types of cancer, such as skin, breast, and lung cancer (Midasala et al., 2024; Kaka et al., 2022; Quanyang et al., 2024). In other medical fields, AI has contributed significantly to improving outcomes, such as in the detection of diabetes in patients (Wu, 2024).

However, AI models are not limited to using a single type of input, such as images; they can also process text as a source of information. Natural language processing (NLP) techniques help extract meaningful information from different types of texts. Among the goals of NLP is predicting whether a person suffers from a particular illness. This approach has been applied, for example, in using AI to predict whether a patient has autism spectrum disorder (ASD), achieving nearly 90% accuracy (Rubio-Martín et al., 2024). Another study related to AI and psychiatry involved the classification of texts about eating disorders (ED) into four categories—texts written by someone with ED, texts that promote ED, informative texts, and scientific texts—achieving nearly 87% accuracy in one of the categories (Benítez-Andrades et al., 2022).

Delving specifically into the convergence between psychiatry and AI, several studies have attempted to assist in the diagnosis or classification of complex mental disorders, such as schizophrenia, depression, or anxiety disorders, using AI (Kodipalli and Devi, 2023; Cortes-Briones et al., 2022; ALSAGRI and YKHLEF, 2020; Nemesure et al., 2021). As shown, applying NLP techniques can help extract relevant information from unstructured data, such as EHRs. The use of EHRs as input for AI has led to the development of models capable of predicting depression crises in patients (Msosa et al., 2023).

In recent years, these efforts have been further expanded in multiple directions. For instance, the prediction of anxiety symptoms in social anxiety disorder has been achieved using multimodal data collected during virtual reality sessions (Park et al., 2025). In another line of work, deep learning models have been developed that outperform clinicians in identifying violence risk from emergency department notes (Dobbins et al., 2024). Transformer-based models have also been employed to detect personal and family history of suicidal ideation in EHRs, yielding F1-scores above 0.90 (Adekkanattu et al., 2024). Furthermore, suicide risk has been phenotyped using multi-label classification strategies based on psychiatric clinical notes (Li et al., 2024).

One of the most challenging scenarios in AI-driven classification involves EHRs, where patients are diagnosed with various mental disorders that share overlapping symptoms. The differentiation between anxiety disorders (ICD-10 F41) and adjustment disorders (ICD-10 F43) is key in the clinical diagnosis and appropriate treatment of patients. Both disorders can present anxious symptoms, but these play a different role in each case. In anxiety disorders (F41), anxious symptoms are central and form part of the core clinical picture. Examples of these disorders include generalized anxiety disorder and social anxiety disorder. Anxiety in these cases does not require a specific external triggering event; that is, the person may experience excessive and ongoing worries about various aspects of life without a clear precipitating factor (World Health Organization, 2019).

On the other hand, adjustment disorders (F43) are characterized by the presence of an identifiable life event or stressor that triggers the symptoms, which may include anxiety, depression, or behavioral changes. These symptoms are a disproportionate psychological response to a stressful situation, such as the loss of a loved one, divorce, or work-related difficulties, and they are time-limited. Unlike anxiety disorders, symptoms in adjustment disorders tend to resolve when the individual adjusts to the new situation or the stressful event is resolved.

While anxiety disorders present anxious symptoms as a central element and do not rely on a clear external trigger, adjustment disorders always have an identifiable stressful event that precipitates the symptoms. This differentiation is fundamental to guide both diagnosis and therapeutic decisions. The importance of distinguishing between these two types of disorders is crucial to avoid misdiagnosis, as clinical interventions for each may differ significantly. A misdiagnosis or confusion between the two could lead to inappropriate treatments, negatively affecting the patient's prognosis (Casey and Bailey, 2011).

For classification purposes, we grouped all ICD-10 codes under the F41 category (Other anxiety disorders) into a single "anxiety disorder" class. This includes panic disorder or episodic paroxysmal anxiety (F41.0), generalized anxiety disorder (F41.1), mixed anxiety disorders (F41.3), other specified anxiety disorders (F41.8), and unspecified anxiety disorder (F41.9). Although our approach focuses on analyzing and classifying existing clinical notes rather than intervening during the initial diagnostic process, structuring and interpreting this information has substantial value. Enhanced documentation quality, retrospective clinical audits, improved training datasets for future models, and support for research activities are some of the ways in which structured clinical information can meaningfully contribute to the mental healthcare system without altering the core diagnostic workflows.

Due to the challenges involved in classifying these two mental disorders, this research demonstrates how AI can achieve highly accurate classification of EHRs, specifically aiming to identify patients diagnosed with adjustment disorder or anxiety disorder. Additionally, this manuscript presents several substantial advancements. The key contributions of this research include:

- **Machine learning models:** We trained several machine learning models in pursuit of the best results. To optimize the performance of each model, a hyperparameter tuning process was carried out. The implementation of this tuning process helped to improve the initial results.

- **BERT-based models:** We explored BERT models, testing two separate pretrained versions, each with distinct training datasets and features that influenced their effectiveness in our tasks.

- **Data balancing process:** Although the dataset is sufficiently large to evaluate the metrics of each model, we applied two data balancing techniques, known as Random Oversampling and Synthetic Minority Oversampling Technique (SMOTE). These techniques were used to assess whether increasing the number of samples in the dataset would allow the models to leverage additional characteristics that could improve the classification task.

- **Real medical dataset:** For this research, clinical notes were provided by the 'Complejo Asistencial Universitario de León' (CAULE). This dataset contains electronic health records of patients diagnosed with adjustment and anxiety disorders. The dataset is entirely self-created, giving it unique value and relevance. From its initial design and data collection to its cleaning, preprocessing, and transformation, every step was meticulously handled to align with the goals of this research. By controlling the entire data treatment process, we gained a deep understanding of the dataset's structure, limitations, and potential insights. This level of control allows for highly tailored analyses and more reliable results. Due to the challenges and restrictions in obtaining clinical notes or other patient information, this dataset holds significant scientific value.

The paper is organized as follows: Section 2, 'Material and Methods', provides a detailed description of the methodology applied, including the collection and preprocessing of the dataset. Section 3, 'Experiments and Results', outlines the experiments conducted and presents the findings, along with a comparison of the various models used. Lastly, Section 4, 'Discussion and Conclusions', brings together the discussion and conclusion to create a unified narrative.

## 2 MATERIAL AND METHODS

This section provides a detailed explanation of the methodology implemented throughout the research. Firstly, section 2.1 describes the process followed to obtain the dataset and how it was transformed from unstructured to structured data. Next, section 2.2 presents the models implemented for this research. Additionally, section 2.3 outlines the hardware and software specifications of the computer used for the research.

### 2.1 Dataset collection and classification

All research involving patient information requires time and the ability to overcome several challenges that arise throughout the process. To begin with, patients' EHRs contain highly sensitive information, which must be protected under strict privacy regulations, as mandated by the European Union's General Data Protection Regulation (GDPR) (European Parlament, 2016). Since patient identification is not required for this research, the clinical notes were anonymized to allow the use of EHRs as a valuable information source, not only in the medical field but also in the field of artificial intelligence (Rao et al., 2023).

An ethics committee was convened and granted us permission to use Spanish EHRs as a dataset for research purposes, ensuring that no patient could be identified. This approval was issued by the Research Ethics Committee for Medicinal Products of the Health Areas of León and Bierzo under the identifier 2303. The EHRs consist of clinical notes from the psychiatry unit of CAULE, written entirely in plain text without any structured data. The dataset comprises 12,921 clinical notes, collected between January 11, 2017, and December 31, 2022. All clinical notes were collected from the Psychiatry Emergency Service of the hospital. Each note documents an urgent psychiatric assessment performed during an emergency department visit. These notes are not part of scheduled outpatient consultations or longitudinal inpatient records, but rather correspond to acute episodes requiring immediate attention. Depending on the evaluation, the patient is either discharged (often with referral for outpatient follow-up) or admitted to inpatient care. Therefore, each note is self-contained and not part of a progressive sequence of visits.

This research was supported by professional psychiatrists who assisted in creating structures to organize the information found in the EHRs. Additionally, these experts provided several guidelines for processing the data. The first step in dataset preprocessing was to remove samples or records where the clinical note was either empty or not properly completed.

To avoid including clinical notes that lacked sufficient or valuable information due to their short length, the experts decided not to consider clinical notes shorter than 600 characters. This threshold was not arbitrary but carefully determined, as it was found that many samples under 600 characters lacked the necessary information to begin structuring the data. Moreover, it was calculated that applying this threshold retained almost 95% of the dataset while ensuring that no clinical notes with insufficient information were included, as shown in Figure 1.

Continuing with the preprocessing phase, the first data extracted from the EHRs were the patient's age and gender. To achieve this, regular expressions were used. A preview of the dataset revealed various patterns that allowed for the extraction of most patients' ages. All phrases structured like '20 years old man' and '30 years old woman' among other possibilities, were captured using a complex regular expression.

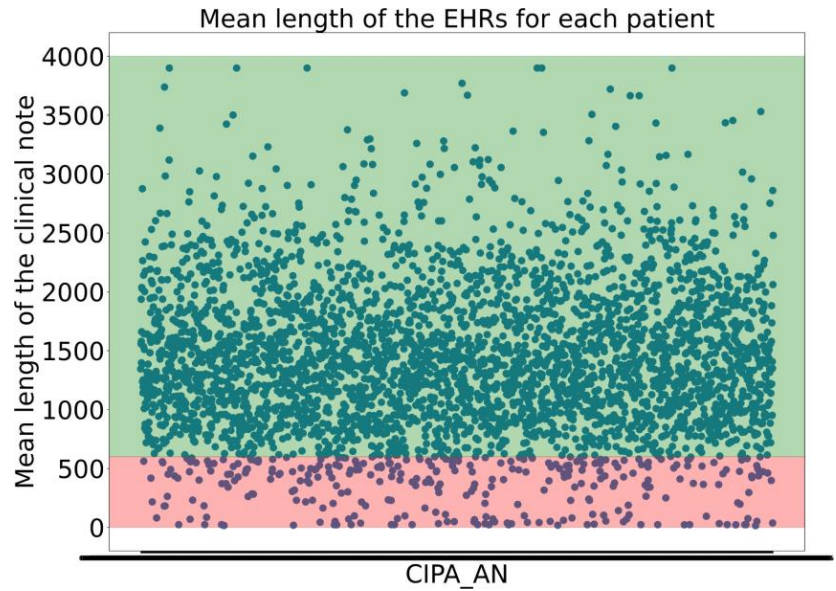

**Figure 1.** Distribution of the mean clinical note length (in characters) per patient. The green area shows that 95% of the patients have an average note length above the 600-character threshold applied. CIPA_AN are the patients ID.

Moreover, the regular expressions were designed to account for common human writing errors, such as missing or extra spaces between words, as well as misspelled words, ensuring the correct extraction of the patient's age. The task of extracting the patient's gender was partially accomplished using the same regular expression, as 'Man' and 'Woman' directly refer to male and female, respectively.

However, in some cases, extracting gender is more challenging, such as in clinical notes where the term 'Patient' is used instead of gender-specific terms like 'Man' or 'Woman'. In these instances, past participle verb forms in Spanish were used to infer the patient's gender. Additionally, when these verbs were absent, marital status indicators like 'single' or 'married', which have gender-specific forms in Spanish, were leveraged to help determine the patient's gender.

The new dataset now consists of several columns. The first column contains the original clinical note. The second column contains the patient's gender, represented as 'V' for male and 'M' for female. The third column records the patient's age. Since the psychiatric clinical notes are plain texts written by professionals summarizing the interview with the patient, the EHRs try to follow the Subjective-Objective-Assessment-Progress (SOAP) standard. However, in this dataset, the information for each section is not clearly delineated, and the majority of notes are composed as unstructured narratives rather than strictly segmented reports.

As a result, identifying the actual diagnosis from these notes is not straightforward. Diagnostic terms such as "anxiety" or "adjustment disorder" may appear in different parts of the note — for instance, in the personal or family history, in symptom descriptions, or as part of comorbidities — without necessarily representing the primary diagnosis. Additionally, anxiety is frequently recorded as a symptom within broader diagnostic categories, adding semantic ambiguity. For these reasons, we did not remove diagnostic terms from the clinical notes during preprocessing. This choice was deliberate, as our aim was to evaluate whether the model could correctly infer the diagnosis based on context, even in the presence of potentially misleading or overlapping terms.

The initial goal of this approach was to extract diagnoses from each clinical note. To achieve this, a Large Language Model (LLM), specifically ChatGPT 4.0, was utilized, as it has proven to be a powerful tool for information extraction in various research studies (Wang et al., 2023). For this research, the ChatGPT API, accessed through Microsoft Azure services, was employed to process 1,000 clinical notes. Prompt engineering techniques, including the use of different roles in API requests (García-Barragán et al., 2024), were applied to enhance the model's performance.

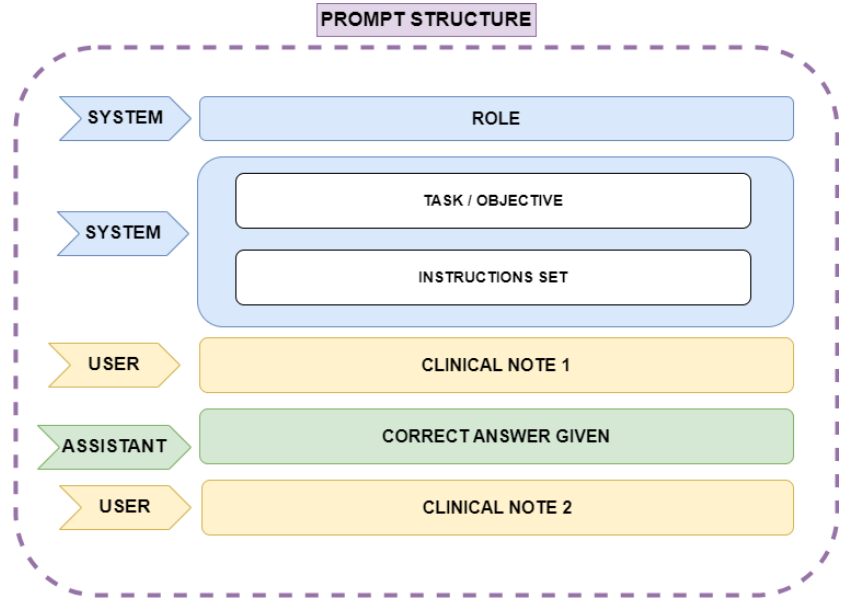

**Figure 2.** Representation of the prompt structure.

One such technique was Few-Shot learning, which involves assigning the model a specific role, explaining the task objective, breaking the task into multiple steps, and providing correct examples of how the task should be performed. This approach ensures that the model understands how to execute the task effectively. Several scientific publications emphasize the value of this Few-Shot prompt engineering technique when working with ChatGPT. In this case, the ChatGPT 4.0 model, which can handle up to 32,000 tokens of conversational context, was used. Since the clinical notes are written in Spanish, the prompt was constructed in Spanish; however, for ease of understanding in this paper, the prompt will be presented in English. The API request format is shown in Figure 2. The prompt structure is explained below:

- **ROLE**: The role assigned to the model. This instruction helps the model adopt an appropriate perspective, focusing on knowledge relevant to the designated role. In this case, the role given was: 'You are an assistant and a linguist specialized in identifying entities within text. You are a leading expert in psychiatry, and I need your help with a very important task in medicine'.

- **TASK and INSTRUCTIONS**: The objective of the task is explained to the model, outlining how it should proceed and detailing how special situations should be handled. Furthermore, the process is broken down into a list of instructions that can be easily followed by the model, as the main problem is split into smaller, manageable tasks.

- **CLINICAL NOTE 1**: Corresponds to the first clinical note provided as plain text.

- **CORRECT ANSWER GIVEN**: A sample of a correct answer is provided to the model for the first clinical note. This example helps the model understand how to proceed. In this case, it was specified that the model should label the diagnosis as 'DX' during entity extraction, using '@@' to indicate the start of the extraction and '##' to indicate the end of the diagnosis extraction. One example of a correct answer given would be 'DX @@ Ansiedad reactiva, Sindrome ansioso-depresivo ##'.

- **CLINICAL NOTE 2**: The next clinical note provided to the model to continue the task.

The final results provided by the LLM were reviewed by experts. After completing the entire preprocessing process, we focused on those clinical notes where patients were diagnosed with Adjustment Disorder or any form of Anxiety Disorder. For this line of research, which centers on these two mental

6/20

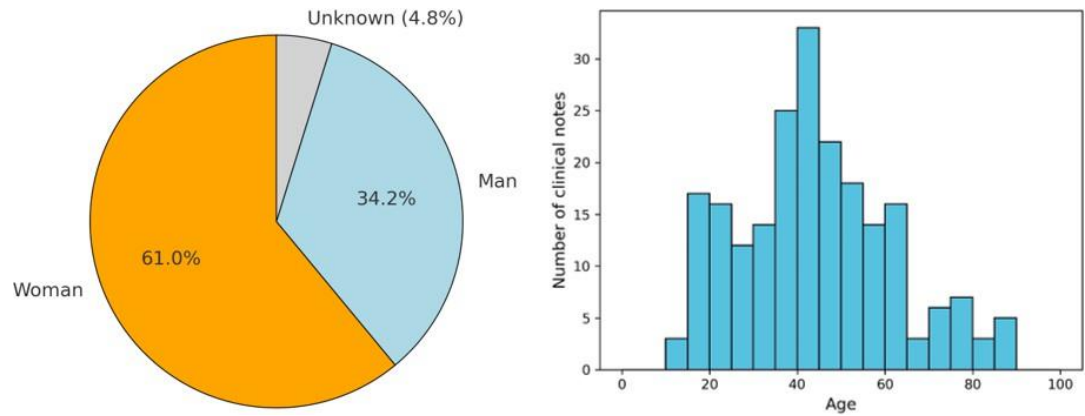

**Figure 3.** Demographic data from the patients found in the clinical notes.

| DX | Man(%) | Unknown(%) | Woman(%) | Mean (Years) | Std (Years) |
|---|---|---|---|---|---|
| Adjustment D. | 32.9 | 6.2 | 60.9 | 44.4 | 18.3 |
| Anxiety D. | 36.6 | 2.4 | 61.0 | 42.7 | 16.4 |

**Table 1.** Percentage distribution by gender and complete age statistics by diagnosis. (Std = Standard Deviation)

disorders, a total of 228 clinical notes were considered: 82 corresponding to Adjustment Disorder and 146 to Anxiety Disorder.

As shown in the left part of Figure 3, of these 228 EHRs, it was found that 61% and 34.2% correspond to clinical notes where the patient is a woman and a man, respectively. Only 4.8% of the notes correspond to cases where the patient's sex is not identified. Additionally, the figure presents the age distribution of patients across the clinical notes, categorized in 5-year intervals. This histogram reveals that the majority of patients fall within the 30 to 50-year age range, with a notable peak around the age of 40. Specifically, the highest number of clinical notes corresponds to patients aged between 35 and 45 years. The distribution also shows that there are fewer clinical notes for patients below 20 years of age and above 70 years, indicating that the majority of the patient population receiving treatment for Adjustment Disorder and Anxiety Disorder tends to be middle-aged. This age trend is consistent with research that shows a high prevalence of these disorders among adults in their working years, likely due to life stressors and social factors often faced during this period (Lotzin et al., 2021).

To complement this general overview, Table 1 presents a detailed demographic breakdown by diagnosis, including gender distribution and descriptive age statistics. In both diagnostic categories, female patients represent the majority: 60.9% in Adjustment Disorder and 61.0% in Anxiety Disorder. Male representation is slightly higher in the Anxiety group (36.6%) compared to Adjustment Disorder (32.9%), while the proportion of patients with unknown gender is relatively low in both groups. Regarding age, the average for patients diagnosed with Adjustment Disorder is 44.4 years (SD = 18.3), and for Anxiety Disorder it is 42.7 years, with a median of 42 years in both cases. These findings confirm that the dataset is predominantly composed of middle-aged individuals, consistent across diagnostic categories, and reinforce the relevance of tailoring classification approaches to this demographic profile.

The age distribution provides valuable demographic insights and helps to contextualize the clinical data being analyzed, especially in terms of tailoring interventions for specific age groups. The relatively lower number of patients in the younger and older age ranges also raises important questions about the underrepresentation of these populations, possibly indicating a need for further exploration of psychiatric care in these demographics.

### 2.2 Machine Learning and Deep Learning models implemented

An intriguing research avenue was explored, focusing on the evaluation of ML and DL models to identify the most accurate approach for addressing the problem. Both linear and non-linear approaches were selected to determine which best suited textual data, given its high dimensionality and potential semantic noise. Below, we describe the theoretical foundation and mathematical formulation of each model, along with the motivation for its selection.

- Random Forest: A versatile and widely used machine learning model that operates by constructing multiple decision trees during training and outputting the mode of the classes or the mean prediction of the individual trees. One key virtue of Random Forest in the context of clinical note classification, is its ability to handle high-dimensional and noisy data effectively, which is common in clinical settings (Al-Showarah et al., 2023). This robustness ensures reliable classification even when dealing with complex medical information as could be the Psychiatry EHRs, improving the model's accuracy and generalization on diverse clinical notes (Góngora Alonso et al., 2022). Mathematically, Random Forest combines multiple decision trees $h_i(x)$, where $h_i(x)$ represents the output of each individual tree $i$, and $N$ is the total number of trees in the forest and the final prediction is obtained through majority voting for classification:

$$\hat{y} = \text{mode}\{h_1(x), h_2(x), \ldots, h_N(x)\} \tag{1}$$

- Support Vector Classifier (SVC): a supervised machine learning model based on the Support Vector Machine (SVM) algorithm. It works by finding the optimal hyperplane that best separates the data into different classes. In this task, SVC tries to maximize the margin between the data points of different classes, which helps in achieving better generalization. SVC can capture complex relationships in the clinical notes, such as the nuanced patterns in clinical language (Elshewey et al., 2023; Lyu et al., 2023). It is also robust to overfitting. The mathematical equation of the decision boundary is:

$$f(x) = w^T x + b \tag{2}$$

where $w$ is the weight vector, $x$ represents the input, and $b$ is the bias term. The optimization process maximizes the margin $\frac{2}{||w||}$ subject to the following constraint where $y_i$ represents the class label:

$$y_i(w^T x_i + b) \geq 1, \quad \forall i \tag{3}$$

- Decision Trees: A type of supervised learning algorithm that makes classifications based on a series of decision rules derived from the input features. The model works by recursively splitting the data into subsets based on feature values, creating branches that represent decision points. Each branch ultimately leads to a leaf node, which represents the predicted class or outcome. This model can help in classifying clinical notes because they are interpretable and can handle both numerical and categorical data. This interpretability is valuable in clinical settings, where understanding the reasoning behind a classification is important for trust and compliance (Vallée et al., 2023). Mathematically, node splitting is based on information gain or Gini index, defined below, where $p_i$ is the proportion of instances of class $i$ in dataset $D$.

$$Gini(D) = 1 - \sum_{i=1}^{C} p_i^2 \tag{4}$$

- XGBoost (eXtreme Gradient Boosting): is a powerful machine learning algorithm that is based on gradient boosting techniques. It works by creating an ensemble of decision trees, where each new tree corrects the errors made by the previous trees. The trees are added sequentially, with each one being optimized to reduce the total error. XGBoost uses gradient descent to minimize a loss function, which allows it to handle complex data patterns effectively (Mir and Sunanda, 2023). It can handle large datasets with high-dimensional features, such as the variety of terms and medical concepts found in clinical text. It also supports regularization, which helps prevent

overfitting—a common issue when working with detailed clinical data. Additionally, XGBoost can efficiently process missing data and is relatively fast, making it well-suited for real-time or large-scale applications in clinical settings (Ulhaq et al., 2023). Each tree in XGBoost is built by minimizing the following regularized loss function, where $l(y_i, \hat{y}_i)$ represents the error function, and $\Omega(f_k)$ penalizes model complexity to prevent overfitting:

$$L(\theta) = \sum_{i=1}^{n} l(y_i, \hat{y}_i) + \sum_{k} \Omega(f_k) \tag{5}$$

- SciBERT (Beltagy et al., 2019): An adapted version of BERT (Bidirectional Encoder Representations from Transformers) that is specifically trained on scientific literature, including biomedical and computer science articles, which makes it well-suited for handling the specialized language in medical contexts. Its transformer-based architecture allows it to understand words in their full context, making it effective for processing clinical notes. When fine-tuned for diagnostic classification, SciBERT can accurately identify patterns in clinical text, recognizing terminology related to various medical conditions (Tang et al., 2023). This makes it particularly valuable for automatically categorizing clinical notes into diagnostic labels, improving the efficiency and accuracy of diagnosis classification tasks in healthcare settings. The transformation function of each layer in BERT-based models is given by the following formula where $Q, K, V$ are the query, key, and value matrices, respectively, and $d_k$ is the key dimension:

$$z_i = \text{softmax}\left(\frac{QK^T}{\sqrt{d_k}}\right) V \tag{6}$$

- DistilBERT (Sanh et al., 2019): is a distilled, or compressed, version of BERT that retains much of BERT's effectiveness while being smaller, faster, and more efficient. It achieves this through a process called 'knowledge distillation', where a smaller model (DistilBERT) is trained to mimic the behavior of a larger model (BERT). DistilBERT has about 40% fewer parameters and is around 60% faster than BERT, but it retains around 97% of BERT's language understanding capabilities. DistilBERT is useful for classifying clinical notes because it provides a good balance between performance and computational efficiency (Abdelhalim et al., 2023). In clinical environments, where there may be constraints on processing power or the need for quick responses, DistilBERT can handle complex language and terminology effectively without requiring the resources that full-sized BERT models do. This makes it suitable large-scale processing of clinical text, where quick and accurate classifications are necessary (Oh et al., 2023; Le et al., 2023).

The selection of models in this study was driven by the need to evaluate both traditional machine learning and deep learning approaches for classifying psychiatric clinical notes. Random Forest and XGBoost were chosen for their strong generalization capabilities, while SVC was included to assess the effectiveness of a linear decision boundary. Decision Tree was selected for its interpretability, which is critical in clinical decision-making. In deep learning, SciBERT was used due to its training on biomedical texts, making it well-suited for clinical language, while DistilBERT was included as a computationally efficient alternative. This diverse set of models ensures a comprehensive evaluation of classification techniques and performance.

### 2.3 Hardware and software Specifications

For the execution of all experiments, Jupyter Notebooks were used. These notebooks were run using Python 3.9 and executed with the following hardware specifications: Intel(R) Core(TM) i7-9700K CPU @ 3.60GHz, 32.0GB RAM, and an NVIDIA GeForce RTX 2080 graphics card. The code used to perform the experiments described in this study is publicly available at the following repository: https://doi.org/10.5281/zenodo.14872650.

# 3 EXPERIMENTS AND RESULTS

## 3.1 Data Preprocessing

Preprocessing clinical notes written in Spanish is a crucial step in preparing data for classification tasks using NLP techniques. Since clinical notes contain unstructured medical information, multiple cleaning and transformation techniques must be applied to enhance data quality and optimize the performance of machine learning models. The following preprocessing steps were performed in detail:

1. **Detection and handling of outliers**: During exploratory analysis, it was detected that clinical notes with fewer than 600 characters rarely contained relevant or substantial information. In many cases, they were administrative records without significant clinical data, and most of them lacked an actual diagnosis. To avoid including non-representative records, a minimum threshold of 600 characters was established in consultation with psychiatrists. By applying this criterion, only notes with sufficient clinical content were processed, ensuring more effective classification.

2. **Handling missing values**: Missing values in patient age and gender were addressed by extracting information from context using regular expressions. This process was carefully designed to improve the statistical reliability of the dataset and ensure a more accurate representation of the patient population.

3. **Lowercasing**: All text was converted to lowercase to reduce unnecessary variability and prevent the model from interpreting words with different casing as distinct entities. In medical language, some words may be capitalized due to writing conventions, but they are semantically equivalent to their lowercase counterparts. This normalization ensured a more homogeneous analysis and reduced the number of unique tokens in the model's vocabulary Chai (2022).

4. **Removal of special characters**: Accents were removed to ensure that a single word is not represented in multiple ways in the model and preserving only spaces and Spanish language characters. RAJESH and HIWARKAR (2023).

5. **Stopword removal**: Frequent words in Spanish that do not contribute meaningful information for classification, such as "el", "de", "que", "en", "un" and "una" were removed using a Spanish stop-word list adapted to the clinical context. Words like "patient," "symptom," or "treatment" were retained, as they are crucial in medical text analysis Sarica and Luo (2021).

6. **Lemmatization**: Lemmatization was applied using the spaCy library to reduce words to their canonical form, decreasing vocabulary dimensionality without losing meaning and maintaining the semantic integrity of clinical notes, which is crucial for understanding the context in medical text Babanejad et al. (2024).

7. **Removal of extra whitespaces**: Removing redundant whitespaces ensures text cleanliness and prevents models from misinterpreting the data as separate entities, improving tokenization accuracy Chai (2022).

This preprocessing pipeline optimized the representation of clinical notes, reducing data noise and improving the model's ability to capture key semantic patterns in medical texts.

## 3.2 Experimental Design

To evaluate the performance of various classification models in distinguishing between diagnoses of Adjustment Disorder and Anxiety Disorder, three different approaches were adopted for handling the training data. These approaches were: (1) without applying any oversampling techniques, (2) using Random Oversampling, and (3) employing the Synthetic Minority Over-sampling Technique (SMOTE).

Oversampling techniques, such as Random Oversampling and SMOTE, are commonly used in machine learning when dealing with imbalanced datasets, where one class is significantly underrepresented compared to the other. In this study, these techniques were explored to see how they could improve the model's ability to correctly classify both disorders, particularly the less common diagnosis, without overfitting to the majority class.

Additionally, to ensure that the distribution of classes (Adjustment Disorder and Anxiety Disorder) remained consistent in both the training and test sets, stratification was applied. Stratification is a method that ensures the class proportions are maintained when splitting the dataset, which is particularly important in imbalanced datasets like this one. Without stratification, there is a risk that one of the sets (training or test) could have a disproportionate number of cases from one class, leading to unreliable performance metrics. By using stratified sampling, both the training (70%) and testing (30%) sets maintain the same distribution of Adjustment Disorder and Anxiety Disorder cases, providing a fair and consistent evaluation during model training and testing.

This step was essential for obtaining reliable performance measurements, as class imbalance can otherwise skew model performance toward the majority class, resulting in misleadingly high accuracy that does not reflect true generalization. Stratification helps prevent this by ensuring that both the minority and majority classes are well-represented in each dataset split, allowing the model to learn from a balanced representation of both diagnoses.

The classification models selected for this task were chosen for their varied approaches and capabilities in handling different types of data. These models included traditional machine learning models such as Random Forest, SVM, and Decision Tree, as well as more advanced models like XGBoost. In addition, two pre-trained transformer-based models, DistilBERT and SciBERT, were employed to leverage their capacity for understanding complex text patterns, particularly in the context of clinical notes.

Each model was evaluated based on two primary metrics: Accuracy and F1-Score. Accuracy provides a general measure of how often the model makes correct predictions and F1-Score gives a more balanced view of model performance in this context.

### 3.3 Results

This subsection describes the results of the experiments conducted on all models, both with and without the use of oversampling techniques. Table 2 presents the performance metrics for each model, highlighting their classification capabilities. The evaluation focuses on key metrics, particularly accuracy and F1-Score, to assess the effectiveness of the models under these conditions.

#### 3.3.1 Models without Oversampling Techniques

The classification models were first evaluated without applying any oversampling techniques. The models demonstrated good performance, though there was significant variability among them.

The XGBoost model achieved the best results, with an accuracy of 96% and an F1-Score of 0.97, indicating excellent classification ability. The Decision Tree model followed, with an accuracy of 93% and an F1-Score of 0.94. These results suggest that tree-based models, particularly XGBoost, are highly effective for the task of classifying clinical notes in this dataset. The Random Forest model also showed satisfactory performance with an accuracy of 81% and an F1-Score of 0.87. However, the SVC model performed worse, with an accuracy of 70% and an F1-Score of 0.81, indicating that it struggled to effectively capture the relationships between features and classes in the data. The pre-trained transformer models (DistilBERT and SciBERT) performed similarly, both achieving an accuracy of 91% and an F1-Score of 0.93. This suggests that these language models, specialized in scientific and clinical text, are particularly useful for this task, outperforming simpler models like SVC and Random Forest. The results obtained without the application of oversampling techniques highlight the strong performance of the XGBoost and Decision Tree models, as well as the effectiveness of pre-trained transformer models. However, the SVC model showed limitations in its classification capability in this context.

| Exp | Metric | Rand.Forest | SVC | Dec.Tree | XGB | DistilBERT | SciBERT |
|---|---|---|---|---|---|---|---|
| WO | Accuracy | 0.81 | 0.70 | 0.93 | **0.96** | 0.91 | 0.91 |
|  | F1-Score | 0.87 | 0.81 | 0.94 | **0.97** | 0.93 | 0.93 |
| RO | Accuracy | 0.81 | 0.70 | 0.87 | **0.96** | 0.55 | 0.91 |
|  | F1-Score | 0.87 | 0.81 | 0.90 | **0.97** | 0.55 | 0.93 |
| SMOTE | Accuracy | 0.87 | 0.70 | 0.88 | **0.96** | 0.91 | 0.91 |
|  | F1-Score | 0.90 | 0.81 | 0.90 | **0.97** | 0.93 | 0.93 |

**Table 2.** Models Performance Across Experiments (Exp = Experiment, XGB = XGBoost, WO = Without Oversampling, RO = Random Oversampling, SMOTE)

### 3.3.2 Models with Random Oversampling

This subsection presents the performance of the models after applying random oversampling to balance the dataset. The introduction of random oversampling had mixed effects on model performance.

The XGBoost model continued to achieve the highest performance, maintaining an accuracy of 96% and an F1-Score of 0.97, consistent with the results without oversampling. This suggests that the XGBoost model is robust to class imbalance, and the oversampling did not significantly alter its ability to classify the clinical notes. The Decision Tree model saw a slight decrease in performance compared to the results without oversampling. Its accuracy dropped from 93% to 87%, and the F1-Score decreased to 0.90. This may suggest that random oversampling introduced some noise, reducing the model's ability to generalize well to the test data. The Random Forest model showed no change in performance, with accuracy and F1-Score remaining at 81% and 0.87, respectively. Similarly, the SVC model's performance remained largely unchanged, with an accuracy of 70% and an F1-Score of 0.81. These results indicate that random oversampling did not provide a substantial improvement for these models in this classification task.

Notably, the DistilBERT model experienced a significant drop in performance when random oversampling was applied. Its accuracy fell to 55%, and its F1-Score dropped to 0.55, suggesting that this transformer-based model was negatively affected by the oversampling technique. On the other hand, SciBERT maintained its strong performance, with an accuracy of 91% and an F1-Score of 0.93, indicating that it was more resilient to the oversampling method. Random oversampling had varying effects on model performance. While it did not lead to improvements in most models, XGBoost maintained its high level of accuracy, and SciBERT remained effective. However, the significant drop in performance for DistilBERT suggests that careful consideration is needed when applying oversampling techniques, especially with transformer-based models.

### 3.3.3 Models with SMOTE

This subsection outlines the performance of the models after applying SMOTE to address class imbalance. Compared to random oversampling, SMOTE generally had a more positive impact on model performance. Once again, the XGBoost model achieved the highest accuracy of 96% and an F1-Score of 0.97, demonstrating consistency across different data balancing techniques. This reinforces XGBoost's robustness and adaptability to imbalanced datasets, as SMOTE did not significantly alter its performance. The Decision Tree model showed a slight improvement with SMOTE compared to random oversampling, reaching an accuracy of 88% and an F1-Score of 0.90. This marginal increase indicates that SMOTE helped the model better generalize, although the performance is still lower than without any oversampling technique. The Random Forest model also saw an improvement, with accuracy rising from 81% to 87% and the F1-Score improving to 0.90. This suggests that SMOTE was more effective than random oversampling in improving the model's ability to classify the minority class without overfitting to the majority class. SVC, however, did not show any noticeable improvement, with its accuracy remaining at 70% and an F1-Score of 0.81, similar to its performance without any oversampling technique. This indicates that SVC's ability to capture relationships in the dataset was not enhanced by SMOTE.

For the transformer-based models, both DistilBERT and SciBERT maintained strong and consistent performance, each achieving an accuracy of 91% and an F1-Score of 0.93. Unlike with random oversampling, DistilBERT's performance remained stable with SMOTE, indicating that the synthetic examples generated by this method may have been better aligned with the underlying data distribution, thereby avoiding the performance degradation observed earlier.

SMOTE had a generally positive impact on model performance, especially for Random Forest and Decision Tree, improving their ability to handle imbalanced data. XGBoost maintained its exceptional performance, and the transformer models continued to show resilience, with DistilBERT recovering from its previous drop in performance with random oversampling.

### 3.4 Hyperparameter Tuning

This subsection presents the results of hyperparameter tuning performed on all models, with and without oversampling techniques, to optimize their performance. The complete hyperparameter search space for each model is summarized in Table 3. This information provides a clearer view of the experimental setup and supports the reproducibility of the results. A 3-fold cross-validation was applied during hyperparameter search to ensure robust evaluation of each configuration. The results for each model after tuning are shown in Table 4. The goal of hyperparameter tuning was to improve the classification metrics, primarily focusing on accuracy and F1-Score.

| Model | Hyperparameters |
|---|---|
| Random Forest | n_estimators: [30, 97, 165, 232, 300]; max_features: ['sqrt', 'log2']; max_depth: [10, 20, 30, 40, 50, None]; min_samples_split: [2, 5, 10]; min_samples_leaf: [1, 2, 4]; bootstrap: [True, False] |
| SVM | C: [0.01, 0.1, 1, 2, 3, 4, 5, 10, 15, 50, 100, 1000]; gamma: [1, 0.1, 0.01, 0.001]; kernel: ['rbf', 'linear', 'sigmoid', 'poly'] |
| Decision Tree | criterion: ['gini', 'entropy', 'log_loss']; splitter: ['best', 'random']; max_depth: 1–29; min_samples_split: 1–19; min_samples_leaf: 1–19; max_features: ['sqrt', 'log2', None]; min_weight_fraction_leaf: [0.0]; random_state: [100] |
| XGBoost | objective: ['binary:logistic', 'binary:logitraw', 'binary:hinge']; learning_rate: [0.1, 0.3, 0.5]; n_estimators: [100, 200, 300, 400]; min_child_weight: [1, 5, 10]; gamma: [1, 2, 5]; subsample: [0.6, 0.8, 1.0]; colsample_bytree: [0.6, 0.8, 1.0]; max_depth: [2, 3, 4, 5] |
| DistilBERT | learning_rate: [1e-5, 3e-5, 5e-5]; batch_size: [8, 16, 32]; epochs: [3, 5, 10] |
| SciBERT | learning_rate: [1e-5, 3e-5, 5e-5]; batch_size: [8, 16]; epochs: [3, 5, 10] |

**Table 3.** Hyperparameter search space for each model.

### 3.4.1 Hyperparameter Tuning without Oversampling

After tuning the hyperparameters, most models showed improved performance when no oversampling techniques were applied. Notably, the Decision Tree model experienced a significant boost, with accuracy rising from 93% to 96% and the F1-Score reaching 0.97. This suggests that fine-tuning the model parameters helped improve its capacity to better distinguish between the classes. The SVC model also demonstrated substantial improvements, with its accuracy increasing from 70% to 88% and its F1-Score reaching 0.91. These improvements reflect the positive impact of hyperparameter optimization on SVC's ability to better handle the complex relationships in the dataset. The Random Forest model improved slightly, with accuracy reaching 86% and an F1-Score of 0.90. Meanwhile, the XGBoost model saw a small decline in accuracy (from 96% to 93%) after hyperparameter tuning, though it still maintained exceptional performance. The slight decrease in performance might indicate that the default parameters were already close to optimal for this model. For the transformer models, both DistilBERT and SciBERT improved their accuracy to 96% and their F1-Scores to 0.97. These gains suggest that tuning transformer-specific parameters, such as learning rate and number of epochs, helped these models better capture the nuances in the clinical text, further boosting their effectiveness.

### 3.4.2 Hyperparameter Tuning with Random Oversampling

In the models trained with random oversampling, hyperparameter tuning led to notable improvements for the SVC model, which saw its accuracy rise to 88% and its F1-Score improve to 0.91, making it much more competitive compared to its previous performance. The Decision Tree model also benefited from tuning, achieving a significant boost in accuracy (96%) and F1-Score (0.97), indicating that the optimized parameters helped counterbalance the challenges posed by the oversampled data. The Random Forest model experienced a slight increase in performance after tuning, with accuracy reaching to 84% and the F1-Score to 0.89. XGBoost maintained its top performance, with both accuracy and F1-Score remaining at 96% and 97% respectively, further emphasizing its robustness to both data imbalance and parameter adjustments.

| Exp | Metric | Rand.Forest | SVC | Dec.Tree | XGB | DistilBERT | SciBERT |
|---|---|---|---|---|---|---|---|
| WO | Accuracy | 0.86 | 0.88 | **0.96** | 0.93 | **0.96** | **0.96** |
|  | F1-Score | 0.90 | 0.91 | **0.97** | 0.94 | **0.97** | **0.97** |
| RO | Accuracy | 0.84 | 0.88 | **0.96** | 0.96 | 0.94 | 0.94 |
|  | F1-Score | 0.89 | 0.91 | **0.97** | 0.97 | 0.95 | 0.95 |
| SMOTE | Accuracy | 0.83 | 0.88 | 0.91 | **0.96** | **0.96** | **0.96** |
|  | F1-Score | 0.87 | 0.91 | 0.94 | **0.97** | **0.97** | **0.97** |

**Table 4.** Models Performance Across Experiments using Hyperparameter Tuning (Exp = Experiment, XGB = XGBoost, WO = Without Oversampling, RO = Random Oversampling, SMOTE)

The transformer models, DistilBERT and SciBERT, both showed improvements with accuracy and F1-Scores rising to 94% and 0.95, respectively, indicating that the combination of random oversampling and tuning positively impacted their ability to classify the clinical notes accurately.

### 3.4.3 Hyperparameter Tuning with SMOTE

When SMOTE was used in conjunction with hyperparameter tuning, the results were similarly positive. The SVC model achieved an accuracy of 88% and an F1-Score of 0.91, consistent with its performance under other oversampling techniques. The Decision Tree model experienced a performance increase, with accuracy rising to 91% and an F1-Score of 0.94. Random Forest, however, showed a slight decrease in performance after tuning, with accuracy dropping to 83% and an F1-Score of 0.87, suggesting that tuning in combination with synthetic data did not favor this model. XGBoost continued to achieve excellent results, maintaining an accuracy of 96% and an F1-Score of 0.97. The transformer models, DistilBERT and SciBERT, also improved after tuning, with both achieving an accuracy of 96% and a F1-Score of 0.97. Hyperparameter tuning was generally effective in enhancing model performance across various techniques. The Decision Tree and SVC models saw the most significant improvements, while XGBoost remained highly consistent. Transformer models also benefited notably from the optimization process.

### 3.5 Computational Performance

The computational time required for hyperparameter tuning varied significantly across the models, as summarized in Table 5. Traditional machine learning models such as Random Forest, SVM, Decision Tree, and XGB exhibited relatively low computational costs, with average times of 0.912 seconds (Random Forest), 0.091 seconds (SVM), 0.003 seconds (Decision Tree), and 0.103 seconds (XGB) per configuration.

In contrast, transformer-based models such as DistilBERT and SciBERT required substantially higher computational resources, with average tuning times of 75.70 seconds and 65.52 seconds per configuration, respectively. These results highlight a clear computational trade-off: while traditional models are significantly more efficient in hyperparameter tuning, transformer-based models demand considerably more processing time. However, prior studies suggest that this increased computational cost often translates into superior performance in terms of accuracy and generalization Benítez-Andrades et al. (2022); Meléndez et al. (2024).

## 4 CONCLUSION AND DISCUSSION

This research has contributed to the field of clinical text classification by examining the effectiveness of different machine learning models in distinguishing between patients diagnosed with Adjustment Disorder and Anxiety Disorder based on clinical notes. Several important findings emerged from this study, highlighting the strengths and limitations of the models employed, as well as the impact of applying oversampling techniques to address class imbalance in the dataset.

| Model | Combinations | Time (s) | Time/Combination |
|---|---|---|---|
| Random Forest | 1080 | 985.267 | 0.912 |
| SVM | 192 | 17.482 | 0.091 |
| Decision Tree | 188442 | 651.747 | 0.003 |
| XGB | 11664 | 1204.597 | 0.103 |
| DistilBERT | 27 | 2043.965 | 75.70 |
| SciBERT | 18 | 1179.506 | 65.52 |

**Table 5.** Computational performance of each model used.

### 4.1 Model Performance

Among the models tested, XGBoost emerged as the best-performing algorithm, consistently demonstrating high accuracy and F1-Score across all experimental setups. Specifically, XGBoost achieved an F1-Score of 0.97 with and without the use of oversampling techniques, proving its robustness in handling the complexities of clinical text classification. The model maintained strong performance even after hyperparameter tuning, confirming its ability to effectively capture class distinctions while maintaining generalization, despite class imbalance in the dataset.

In contrast, the Support Vector Classifier (SVC) model exhibited the weakest performance, particularly without oversampling, where it struggled with an accuracy of 70% and an F1-Score of 0.81. This is likely due to the sensitivity of SVC to imbalanced datasets, where the minority class may be overshadowed by the majority class. Although hyperparameter tuning and oversampling techniques such as Random Oversampling and SMOTE improved SVC's performance (raising the F1-Score to 0.91 in some cases), its results remained below those of more advanced models like XGBoost, SciBERT, and DistilBERT. These findings indicate that while SVC can be a reliable option in certain domains, it may not be well-suited for imbalanced clinical text classification tasks without significant adjustments. Figures 4 and 5 illustrate the comparative performance of the machine learning and deep learning models, respectively.

### 4.2 The Impact of Oversampling Techniques

A key aspect of this research was the evaluation of two oversampling techniques: Random Oversampling and SMOTE. The results indicate that oversampling had a varying impact on model performance, particularly for models sensitive to class imbalance.

For models like Random Forest and XGBoost, Random Oversampling did not result in significant performance gains, and in some cases, even led to a slight drop in performance. For instance, DistilBERT experienced a considerable decline when random oversampling was applied, with the F1-Score dropping to 0.55. This suggests that Random Oversampling may introduce noise, particularly in more complex models, and thus does not consistently benefit all models.

SMOTE, on the other hand, proved to be a more effective technique for improving performance across various models. In particular, SMOTE enhanced the performance of models like Decision Tree and Random Forest, which achieved F1-Scores of 0.90 and 0.90, respectively, when applied. Furthermore, models such as XGBoost and transformer-based models like DistilBERT and SciBERT maintained their strong performance with SMOTE, both achieving F1-Scores of 0.97. The results indicate that SMOTE helped these models create more balanced decision boundaries without duplicating existing data points, leading to more robust classification outcomes.

It was found that, while oversampling techniques generally improved performance, SMOTE was more effective across a range of models, particularly for complex architectures like XGBoost and transformer-based models.

### 4.3 Comparison Between Transformer Models and Traditional Machine Learning

The transformer-based models, DistilBERT and SciBERT, demonstrated strong results throughout the experiments, confirming their potential for natural language processing tasks in the healthcare domain. In comparison to traditional machine learning models such as Random Forest and SVC, the transformers-based models were better at capturing the nuances of clinical language, particularly when no oversampling techniques were applied.

SciBERT, pretrained on scientific texts, was particularly noteworthy, achieving an F1-Score of 0.97 with SMOTE, highlighting its strength in parsing and classifying the specialized terminology found in clinical notes. DistilBERT performed well across all setups, reinforcing the potential of transformer-based models in healthcare text classification tasks. Notably, these transformer models remained highly effective, even when class imbalance was not addressed through oversampling techniques.

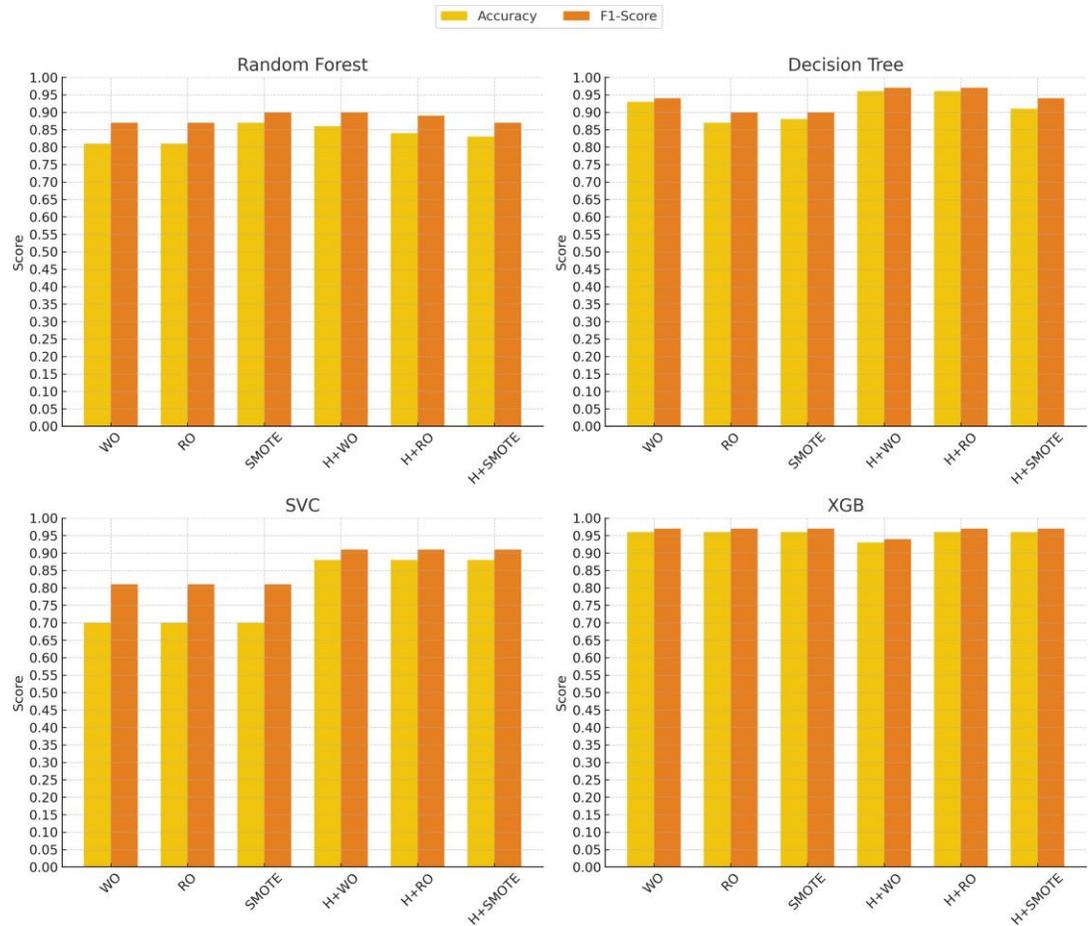

**Figure 4.** Visual comparison of accuracy and F1-score across machine learning models using different configurations (WO = Without Oversampling, RO = Random Oversampling, H = Hyperparameters)

### 4.4 Impact of Hyperparameter Tuning

Hyperparameter tuning was a critical component in this study, as it helped optimize the performance of all the machine learning models. The results clearly show that hyperparameter tuning had a significant impact on improving classification metrics, particularly for models that initially struggled with imbalanced data or suboptimal settings.

The most notable improvement was observed in the Decision Tree model, where hyperparameter tuning increased its accuracy from 93% to 96% and its F1-Score to 0.97 when no oversampling was applied. This demonstrates that tuning allowed the Decision Tree model to make better splits and generalize more effectively on the data, leading to performance that matched the top-performing models such as XGBoost.

Similarly, the SVC model benefited substantially from hyperparameter tuning. Initially, SVC struggled with imbalanced data, but after tuning, its accuracy increased to 88% and its F1-Score improved to 0.91. These improvements indicate that carefully optimizing parameters like the kernel and gamma allowed SVC to better distinguish between the diagnostic categories.

The transformer models, DistilBERT and SciBERT, also saw improvements with hyperparameter tuning. Both DistilBERT and SciBERT achieved an accuracy of 96% and an F1-Score of 0.97 after tuning. These results suggest that while the transformers performed well without significant tuning, the fine-tuning of parameters like learning rate and number of epochs still provided marginal performance boosts.

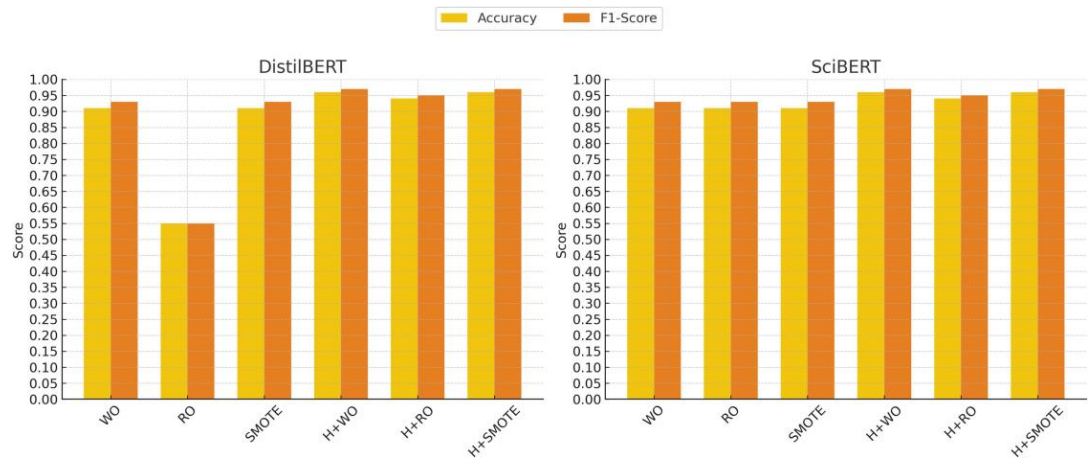

**Figure 5.** Visual comparison of accuracy and F1-score across deep learning models using different configurations (WO = Without Oversampling, RO = Random Oversampling, H = Hyperparameters)

For the XGBoost model, however, hyperparameter tuning led to a slight reduction in accuracy, dropping from 96% to 93%, although its F1-Score remained high at 0.94. This suggests that XGBoost may have already been operating near its optimal settings.

Hyperparameter tuning proved to be a valuable step in improving model performance. While simpler models like Decision Tree and SVC saw the most pronounced benefits, even advanced models such as transformers and XGBoost showed gains in certain metrics, reaffirming the importance of hyperparameter optimization in machine and deep learning workflows.

### 4.5 Limitations and Future Directions

Despite the strong performance of the models tested, several limitations should be acknowledged. First, the dataset, although preprocessed, might still contain noise inherent to clinical notes, such as inconsistent terminology or incomplete information, which could affect model performance. Future studies could focus on refining the preprocessing pipeline to handle these nuances more effectively, potentially leading to further improvements in classification accuracy.

Additionally, while this study demonstrated the value of oversampling techniques, there are alternative methods for addressing class imbalance that were not explored, such as cost-sensitive learning or under-sampling methods, which could be examined in future research. These techniques might offer more efficient solutions, especially in scenarios where oversampling introduces overfitting or data redundancy. Future directions could also consider exploring Variational Autoencoders (VAEs) as a generative approach for oversampling.

Another avenue for future research involves evaluating additional classification models beyond those tested in this study. Exploring more advanced deep learning architectures or novel transformer-based models could further enhance classification performance, particularly in complex diagnostic scenarios. Moreover, expanding the dataset to include clinical notes from patients presenting similar symptomatology but ultimately receiving different diagnoses would provide a more challenging and realistic classification setting. This would help assess the models' ability to capture subtle clinical distinctions, which is critical in psychiatric evaluation.

Finally, although transformer models performed well, their computational cost and the need for large datasets for fine-tuning present practical challenges. Future work could explore the use of more efficient transformer architectures or hybrid models that combine the strengths of transformers and traditional machine learning approaches.

# REFERENCES

Abdelhalim, N., Abdelhalim, I., and Batista-Navarro, R. (2023). Training models on oversampled data and a novel multi-class annotation scheme for dementia detection. In Naumann, T., Ben Abacha, A., Bethard, S., Roberts, K., and Rumshisky, A., editors, *Proceedings of the 5th Clinical Natural Language Processing Workshop*, pages 118–124, Toronto, Canada. Association for Computational Linguistics.

Adekkanattu, P., Furmanchuk, A., Wu, Y., Pathak, A., Patra, B. G., Bost, S., Morrow, D., Wang, G. H.-M., Yang, Y., Forrest, N. J., Luo, Y., Walunas, T. L., Lo-Ciganic, W., Gelad, W., Bian, J., Bao, Y., Weiner, M., Oslin, D., and Pathak, J. (2024). Deep learning for identifying personal and family history of suicidal thoughts and behaviors from EHRs. *npj Digital Medicine*, 7(1):260.

Al-Showarah, S., Al-Taie, A., Salman, H. E., Alzyadat, W., and Alkhalaileh, M. (2023). Predicting quality medical drug data towards meaningful data using machine learning. *International Journal of Advanced Computer Science and Applications*, 14(8).

Alanazi, A. (2023). Clinicians' Views on Using Artificial Intelligence in Healthcare: Opportunities, Challenges, and Beyond. *Cureus*, 15(9):e45255. Place: United States.

ALSAGRI, H. S. and YKHLEF, M. (2020). Machine learning-based approach for depression detection in twitter using content and activity features. *IEICE Transactions on Information and Systems*, E103.D(8):1825–1832.

Babanejad, N., Davoudi, H., Agrawal, A., An, A., and Papagelis, M. (2024). The role of preprocessing for word representation learning in affective tasks. *IEEE Transactions on Affective Computing*, 15(1):254–272.

Beltagy, I., Lo, K., and Cohan, A. (2019). SciBERT: A pretrained language model for scientific text. In Inui, K., Jiang, J., Ng, V., and Wan, X., editors, *Proceedings of the 2019 Conference on Empirical Methods in Natural Language Processing and the 9th International Joint Conference on Natural Language Processing (EMNLP-IJCNLP)*, pages 3615–3620, Hong Kong, China. Association for Computational Linguistics.

Benítez-Andrades, J. A., Alija-Pérez, J.-M., Vidal, M.-E., Pastor-Vargas, R., and García-Ordás, M. T. (2022). Traditional machine learning models and bidirectional encoder representations from transformer (bert)–based automatic classification of tweets about eating disorders: Algorithm development and validation study. *JMIR Med Inform*, 10(2):e34492.

Briganti, G. and Le Moine, O. (2020). Artificial intelligence in medicine: Today and tomorrow. *Frontiers in Medicine*, 7.

Casey, P. and Bailey, S. (2011). Adjustment disorders: the state of the art. *World Psychiatry*, 10(1):11–18.

Chai, C. P. (2022). Comparison of text preprocessing methods. *Natural Language Engineering*, 29:509 – 553.

Cortes-Briones, J. A., Tapia-Rivas, N. I., D'Souza, D. C., and Estevez, P. A. (2022). Going deep into schizophrenia with artificial intelligence. *Schizophrenia Research*, 245:122–140. Computational Approaches to Understanding Psychosis.

Dobbins, N. J., Chipkin, J., Byrne, T., Ghabra, O., Siar, J., Sauder, M., Huijon, R. M., and Black, T. M. (2024). Deep learning models can predict violence and threats against healthcare providers using clinical notes. *npj Mental Health Research*, 3(1):61.

Elshewey, A. M., Shams, M. Y., El-Rashidy, N., Elhady, A. M., Shohieb, S. M., and Tarek, Z. (2023). Bayesian optimization with support vector machine model for parkinson disease classification. *Sensors*, 23(4).

European Parlament (2016). Regulation (eu) 2016/679 of the european parliament and of the council of 27 april 2016 on the protection of natural persons with regard to the processing of personal data and on the free movement of such data, and repealing directive 95/46/ec (general data protection regulation).

García-Barragán, A., González Calatayud, A., Solarte-Pabón, O., Provencio, M., Menasalvas, E., and Robles, V. (2024). GPT for medical entity recognition in Spanish. *Multimedia Tools and Applications*.

Góngora Alonso, S., Marques, G., Agarwal, D., De la Torre Díez, I., and Franco-Martín, M. (2022). Comparison of machine learning algorithms in the prediction of hospitalized patients with schizophrenia. *Sensors*, 22(7).

Henriques, G. and Michalski, J. (2020). Defining behavior and its relationship to the science of psychology. *Integrative Psychological and Behavioral Science*, 54(2):328–353.

Kaka, H., Michalopoulos, G., Subendran, S., Decker, K., Lambert, P., Pitz, M., Singh, H., and Chen, H. (2022). Pretrained Neural Networks Accurately Identify Cancer Recurrence in Medical Record.

*Studies in health technology and informatics*, 294:93–97. Place: Netherlands.

Kendler, K. S., Zachar, P., and Craver, C. (2011). What kinds of things are psychiatric disorders? *Psychological Medicine*, 41(6):1143–1150.

Kodipalli, A. and Devi, S. (2023). Analysis of fuzzy based intelligent health care application system for the diagnosis of mental health in women with ovarian cancer using computational models. *Intelligent Decision Technologies*, 17(1):31–42. Publisher: IOS Press.

Le, T.-D., Jouvet, P., and Noumeir, R. (2023). A small-scale switch transformer and nlp-based model for clinical narratives classification. *ArXiv*, abs/2303.12892.

Li, Z., Hu, Y., Lane, S., Selek, S., Shahani, L., Machado-Vieira, R., Soares, J., Xu, H., Liu, H., and Huang, M. (2024). Suicide phenotyping from clinical notes in safety-net psychiatric hospital using multi-label classification with pre-trained language models.

Liu, R., Rong, Y., and Peng, Z. (2020). A review of medical artificial intelligence. *Global Health Journal*, 4(2):42–45.

Lotzin, A., Krause, L., Acquarini, E., Ajdukovic´, D., Ardino, V., Arnberg, F., Bo¨ttche, M., Bragesjo¨, M., Dragan, M., Figueiredo-Braga, M., Gelezelyte, O., Grajewski, P., Anastassiou-Hadjicharalambous, X., Javakhishvili, J., Kazlauskas, E., Lenferink, L., Lioupi, C., Lueger-Schuster, B., Tsiskarishvili, L., Mooren, T., Sales, L., Stevanovic´, A., Zrnic´, I., Scha¨fer, I., and Consortium, A. S. (2021). Risk and protective factors, stressors, and symptoms of adjustment disorder during the covid-19 pandemic – first results of the estss covid-19 pan-european adjust study. *European Journal of Psychotraumatology*, 12.

Lyu, Y., Xu, Q., Yang, Z., and Liu, J. (2023). Prediction of patient choice tendency in medical decision-making based on machine learning algorithm. *Frontiers in Public Health*, 11.

Mele´ndez, R., Ptaszynski, M., and Masui, F. (2024). Comparative investigation of traditional machine-learning models and transformer models for phishing email detection. *Electronics*, 13(24).

Midasala, V. D., Prabhakar, B., Krishna Chaitanya, J., Sirnivas, K., Eshwar, D., and Kumar, P. M. (2024). Mfeuslnet: Skin cancer detection and classification using integrated ai with multilevel feature extraction-based unsupervised learning. *Engineering Science and Technology, an International Journal*, 51:101632.

Mir, S. and Sunanda (2023). "heart disease prediction and severity level classification": A machine learning approach with feature selection technique. *2023 14th International Conference on Computing Communication and Networking Technologies (ICCCNT)*, pages 1–7.

Moitra, M., Santomauro, D., Collins, P. Y., Vos, T., Whiteford, H., Saxena, S., and Ferrari, A. J. (2022). The global gap in treatment coverage for major depressive disorder in 84 countries from 2000–2019: A systematic review and bayesian meta-regression analysis. *PLOS Medicine*, 19(2):1–16.

Msosa, Y. J., Grauslys, A., Zhou, Y., Wang, T., Buchan, I., Langan, P., Foster, S., Walker, M., Pearson, M., Folarin, A., Roberts, A., Maskell, S., Dobson, R., Kullu, C., and Kehoe, D. (2023). Trustworthy data and ai environments for clinical prediction: Application to crisis-risk in people with depression. *IEEE Journal of Biomedical and Health Informatics*, 27(11):5588–5598.

Nemesure, M. D., Heinz, M. V., Huang, R., and Jacobson, N. C. (2021). Predictive modeling of depression and anxiety using electronic health records and a novel machine learning approach with artificial intelligence. *Scientific Reports*, 11(1):1980.

Nimavat, N., Hasan, M. M., Mandala, G., Singh, S., Bhangu, R., and Bibi, S. (2023). Mortality Rate in Schizophrenia. In Chatterjee, I., editor, *Cognizance of Schizophrenia:: A Profound Insight into the Psyche*, pages 303–312. Springer Nature Singapore, Singapore.

Oh, J., Kim, M., Park, H., and Oh, H. (2023). Are you depressed? analyze user utterances to detect depressive emotions using distilbert. *Applied Sciences*.

Park, J.-H., Shin, Y.-B., Jung, D., Hur, J.-W., Pack, S. P., Lee, H.-J., Lee, H., and Cho, C.-H. (2025). Machine learning prediction of anxiety symptoms in social anxiety disorder: utilizing multimodal data from virtual reality sessions. *Frontiers in Psychiatry*, 15.

Quanyang, W., Yao, H., Sicong, W., Linlin, Q., Zewei, Z., Donghui, H., Hongjia, L., and Shijun, Z. (2024). Artificial intelligence in lung cancer screening: Detection, classification, prediction, and prognosis. *Cancer Medicine*, 13(7):e7140. e7140 CAM4-2023-11-5724.R1.

RAJESH, M. A. and HIWARKAR, D. T. (2023). Exploring preprocessing techniques for natural languagetext: A comprehensive study using python code. *international journal of engineering technology and management sciences*.

Rao, K. N., Arora, R. D., Dange, P., and Nagarkar, N. M. (2023). NLP AI Models for Optimizing Medical

Research: Demystifying the Concerns. *Indian Journal of Surgical Oncology*, 14(4):854–858.

Rubio-Martín, S., García-Ordás, M. T., Bayón-Gutiérrez, M., Prieto-Fernández, N., and Benítez-Andrades, J. A. (2024). Enhancing ASD detection accuracy: a combined approach of machine learning and deep learning models with natural language processing. *Health Information Science and Systems*, 12(1):20.

Sanh, V., Debut, L., Chaumond, J., and Wolf, T. (2019). Distilbert, a distilled version of bert: smaller, faster, cheaper and lighter. *ArXiv*, abs/1910.01108.

Sarica, S. and Luo, J. (2021). Stopwords in technical language processing. *PLOS ONE*, 16(8):1–13.

Tang, X., Tran, A., Tan, J., and Gerstein, M. B. (2023). Gersteinlab at mediqa-chat 2023: Clinical note summarization from doctor-patient conversations through fine-tuning and in-context learning. *arXiv preprint*, 2305:546–554.

Ulhaq, S., Khan, G. Z., Ulhaq, I., Ullah, I., Rabbi, F., Zaman, G., and Khan (2023). Epilepsy seizures classification with eeg signals: A machine learning approach. *Journal of Computer Science and Technology Studies*.

Vallée, R., Vallée, J., Guillevin, C., Lallouette, A., Thomas, C., Rittano, G., Wager, M., Guillevin, R., and Vallée, A. (2023). Machine learning decision tree models for multiclass classification of common malignant brain tumors using perfusion and spectroscopy mri data. *Frontiers in Oncology*, 13.

Wang, S., Sun, X., Li, X., Ouyang, R., Wu, F., Zhang, T., Li, J., and Wang, G. (2023). Gpt-ner: Named entity recognition via large language models.

World Health Organization (2019). International statistical classification of diseases and related health problems 10th revision (icd-10).

World Health Organization (2021). *Mental health atlas 2020.* World Health Organization.

World Health Organization (2022a). Mental disorders.

World Health Organization (2022b). Mental health and covid-19: Early evidence of the pandemic's impact: Scientific brief.

Wu, K. (2024). Optimizing diabetes prediction with machine learning: Model comparisons and insights. *Journal of Science & Technology*, 5(4):41–51.

Zhang, F., Geng, J., Zhang, D.-G., Gui, J., and Su, R. (2023). Prediction of cancer recurrence based on compact graphs of whole slide images. *Computers in Biology and Medicine*, 167:107663.